\documentclass[journal]{IEEEtran}
\ifCLASSINFOpdf
\else
\fi

\usepackage{cite}
\usepackage{amsmath}
\usepackage{algorithm}
\usepackage{algorithmic}
\usepackage{array}
\usepackage{multirow}
\usepackage{booktabs}
\usepackage{graphicx}
\usepackage{amssymb}
\usepackage{subfigure}

\usepackage[caption=false,font=footnotesize]{subfig}
\usepackage{stfloats}
\usepackage{url}
\usepackage{times}
\usepackage{makecell}

\begin{document}

\title{LF-YOLO: A Lighter and Faster YOLO for Weld Defect Detection of X-ray Image}

\author{Moyun Liu, Youping Chen, Lei He, Yang Zhang, Jingming Xie
\thanks{\emph{(Corresponding author: Jingming Xie.)}}
\thanks{M. Liu, Y. Chen, L. He and J. Xie are with the School of Mechanical Science and Engineering,
Huazhong University of Science and Technology, Wuhan 430074, China
(e-mail: lmomoy@hust.edu.cn; ypchen@hust.edu.cn; leihe\_01@hust.edu.cn; xjmhust@hust.edu.cn;)}
\thanks{Y. Zhang is with the School of Mechanical Engineering, Hubei University of
Technology, Wuhan 430068, China, and also with the National Key
Laboratory for Novel Software Technology, Nanjing University, Nanjing 210023,
China (e-mail: yzhangcst@smail.nju.edu.cn)}% <-this % stops a space
%\thanks{Color versions of one or more of the figures in this article are
%available online at http://ieeexplore.ieee.org.}
%\thanks{Digital Object Identifier}
}

% The paper headers
\markboth{Journal of \LaTeX\ Class Files,~Vol.~14, No.~8, August~2015}%
{Shell \MakeLowercase{\textit{et al.}}: Bare Demo of IEEEtran.cls for IEEE Journals}

% make the title area
\maketitle

\begin{abstract}
X-ray image plays an important role in manufacturing industry for quality assurance, because it can reflect the internal condition of weld region. However, the shape and scale of different defect types vary greatly, which makes it challenging for model to detect weld defects. In this paper, we propose a weld defect detection method based on convolution neural network, namely Lighter and Faster YOLO (LF-YOLO). In particularly, a reinforced multiscale feature (RMF) module is designed to implement both parameter-based and parameter-free multi-scale information extracting operation. RMF enables the extracted feature map capable to represent more plentiful information, which is achieved by superior hierarchical fusion structure. To improve the performance of detection network, we propose an efficient feature extraction (EFE) module. EFE processes input data with extremely low consumption, and improves the practicability of whole network in actual industry. Experimental results show that our weld defect detection network achieves satisfactory balance between performance and consumption, and reaches 92.9 mean average precision (mAP$_{50}$) with 61.5 frames per second (FPS). To further prove the ability of our method, we test it on public dataset MS COCO, and the results show that our LF-YOLO has a outstanding versatility detection performance. The code is available at https://github.com/lmomoy/LF-YOLO.
\end{abstract}


\section{Introduction}
\begin{figure}[!t]
  \centering
  \includegraphics[width=3.2in]{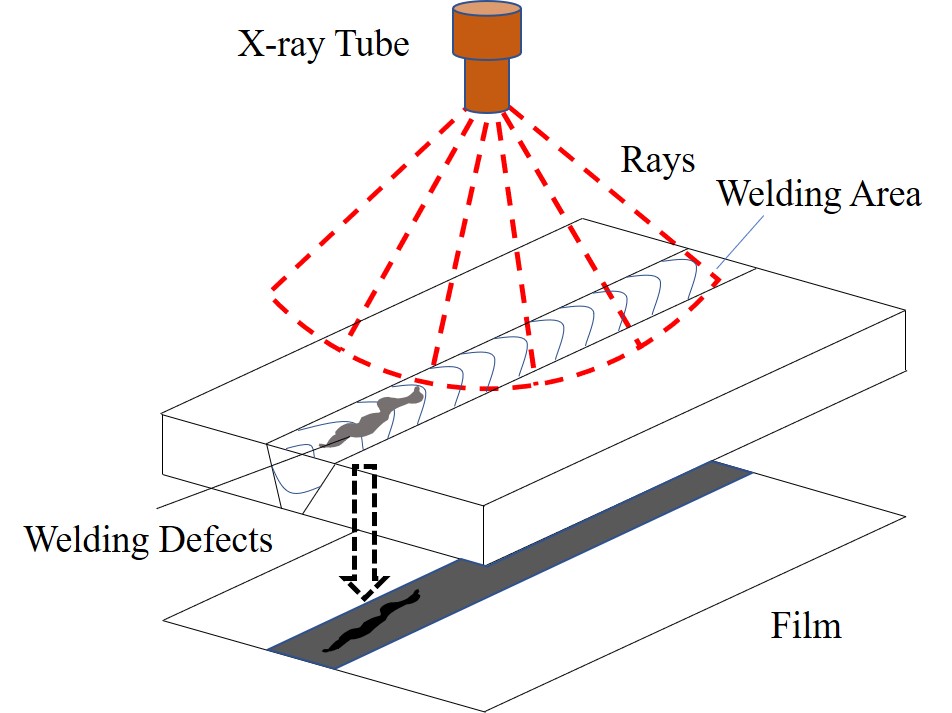}
  \caption{Production process of X-ray weld image. The X-ray tube emits rays to metal, and the welding area and internal defects are revealed on the bottom film.}
  \label{fig:X-ray process}
\end{figure}

Welding technology is common in various production applications, such as aircraft, mechanical equipment and shipbuilding~\cite{yang2021automatic}. Hence, the quality of weld determines the safety of industrial operation~\cite{liu2021lightweight}. However, either manual or robotic welding will inevitably produce weld defects, which is a potential hazard for daily production. To address the problem, people utilize X-ray technology to reflect internal defect of weld into image as shown in Fig.~\ref{fig:X-ray process}, and detect them through expert or computer vision model. Because workers are subjective and easy to fatigue, manual detection is inefficient and unsustainable.

\begin{figure}[!t]
  \centering
  \includegraphics[width=3.2in]{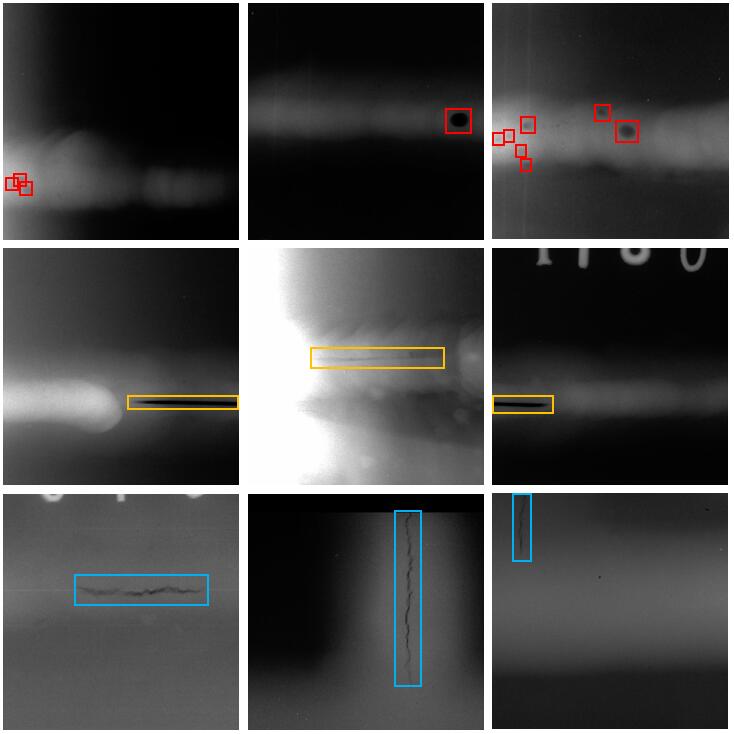}
  \caption{Different weld defect samples. The first row is blow hole (red box), and second row is incomplete peneration (yellow box), while the third row is crack (blue box). The scale of weld defects vary greatly among different defect classes. Meantime, the defect is hard to be recognized due to the blurred boundaries and similar texture between foreground and background.}
  \label{fig:weld}
\end{figure}

The context of weld image is complicated, and there are blurred boundaries and similar texture between defect and background. In addition, the scales and shapes of defects vary greatly among different classes, which can be seen in Fig.~\ref{fig:weld}. The size of blow hole is relatively small, while incomplete peneration and crack differ greatly in aspect ratio. All of these factors bring great challenges to the detection model~\cite{rashwan2020matrixnets}, and it is required to capture abundant contextual information. Specifically, local feature is beneficial to represent the boundary, shape, and geometric texture of defect, while global feature is vital for classification and distinguishing foreground and background.

Convolution neural networks (CNN) is a self-learning model which can extract superior image features automatically based on well-designed structure. The initial CNN models are weak to combine multi-scale features, because of the plain structure of them. To address this problem, many feature fusion methods~\cite{2017Feature}~\cite{yu2015multi}~\cite{bochkovskiy2020yolov4} are proposed to better utilize local and global context simultaneously. In this paper, we classify them into two categories, i.e., parameter-based and parameter-free operations. Parameter-based method introduces new learnable module, and it can encode new representation. Parameter-free method can be regard as a re-utilization based on existing information. It exploits extracted feature map thoroughly and makes the most of them. Weld defect detection has introduced advanced neural network technology, and achieved decent performance. However, these researches usually use simple or classical CNNs directly~\cite{2016Automated}~\cite{2019Weld}~\cite{2020Automatic}, and hardly make any adaptive improvement. As mentioned above, the scale variation and contextual information is vital for weld image, but there are little literatures to solve these problems.~\cite{2020Multi} analyzes the defect features at different scales in the parameter space, but it takes expensive calculations and is inefficient for industry.

In this paper, we propose an reinforced multiscale feature (RMF) module, which combines both of parameter-based and parameter-free operations. RMF module firstly contains a basic parameter-free hierarchical structure, which generates multiple feature maps obtained from maxpool operations of different sizes. Furthermore, within each branch of basic hierarchy, new features are produced through learning potential information implicitly, and the process is parameter-based. Finally, the output data of each hierarchy would be fused for finer estimation. Besides the contribution of multi-scale feature utilization, original feature extraction also determines the performance of the network. To effectively extract feature of weld defect, we design an efficient feature extraction (EFE) module elaborately, and build a superior backbone by stacking EFE repeatedly. Based on the above two core components, a Lighter and Faster YOLO (LF-YOLO) is proposed to detect weld defect of X-ray image.

In summary, this work makes the following contributions.
\begin{itemize}
  \item A novel multi-scale fusion module named RMF is proposed. It can combine local and global cues of X-ray image by using parameter-based and parameter-free methods simultaneously.
  \item To efficiently learn representation, we design a novel EFE module as the unit of backbone, and it can extract meaningful feature with few parameters and low computation.
  \item We propose a LF-YOLO for weld defect detection of X-ray image, which can deal with multiple defect classes, and the proposed network is memory and computation friendly.
  \item We validate the effectiveness of our LF-YOLO according to extensive experiments on our datasets. In addition, we test our LF-YOLO on public object detection dataset to prove its versatility as well.
\end{itemize}

\section{Related works}
\subsection{Object detection}

Convolution neural networks have succeed in object detection, since it owns many advantages such as weight sharing and local connection. Object detection networks are mainly divided into two categories: one-stage and two-stage. The model based on one-stage strategy regresses the classification and location of target directly, which makes it can achieve fast detection speed, and they are high efficiency and hardware-friendly. One-stage detector can be designed based on anchor. YOLO v3~\cite{redmon2018yolov3}, SSD~\cite{liu2016ssd} and RetinaNet~\cite{lin2017focal} place anchor boxes densely over feature maps, then predict object category and anchor box offsets. VFNet~\cite{zhang2020varifocalnet}, RepPoints~\cite{yang2019reppoints} are anchor-free detectors, which predict some important points, such as corner point or center point to form bounding box. Compared with one-stage network, the two-stage methods such as Faster R-CNN~\cite{ren2015faster}, Cascade R-CNN~\cite{cai2018cascade}, R-FCN~\cite{dai2016r} and Dynamic R-CNN~\cite{DynamicRCNN} have a lower inference speed, because they produce region proposal to distinguish foreground and background at first. However, the refinement design improve the detection performance and makes them more suitable for high accuracy scenarios.

Nevertheless, the high computation requirement limits the application of CNNs in actual industry. To alleviate this problem, many lightweight networks are proposed to reduce the complexity of model. SqueezeNet~\cite{iandola2016squeezenet} proposes a novel Fire Module, and it reduces parameters by using $1\times1$ convolution to replace $3\times3$ convolution. The MobileNet~\cite{howard2017mobilenets}~\cite{2018MobileNetV2}~\cite{howard2019searching} series networks achieve model compression mainly relying on depthwise separable convolution. A cheap operation is introduced in GhostNet~\cite{han2020ghostnet}, which has fewer parameters while obtaining the same number of feature maps.

\subsection{Multi-scale feature utilization}
Existing CNN-based algorithms suffer from the problem that the convolutional features are scale-sensitive in object detection task~\cite{8478157}. Many methods attempt to make the network fit different scales by utilizing multi-scale feature, and we divide them into two categories: parameter-based and parameter-free methods.

\textbf{Parameter-based method.} Learnable parameter can exploit and extract implicit feature from different scales. Dilated convolution~\cite{yu2015multi} aggregates multi-scale contextual information through changing the dilation ratio of convolution. ASPP~\cite{chen2017deeplab} operates dilated convolutions with different rates on input feature maps. Then, the extracted features via different branches are concatenated at last to combine the information from different receptive fields. Based on~\cite{chen2017deeplab}, RFBNet~\cite{2017Receptive} further introduces convolutions with different kernel sizes to enhance the scale difference of learned feature.

\textbf{Parameter-free method.} In general, semantic information would be produced in the deeper layers, while the shallower layers mainly include spatial location information~\cite{9346044}. Parameter-free methods mainly aim at making fully use of these existing features without introducing new learnable unit. The simplest way is adding or concatenating feature maps from different layers. Furthermore, FPN~\cite{2017Feature} designs a top-down network structure, and send semantic information of the top layers to the bottom layers. PANet~\cite{2018Path} adds the bottom-up path after the network structure of the FPN, further enriching and enhancing features. Spatial pyramid pooling~\cite{bochkovskiy2020yolov4} is another parameter-free operation, which uses several maxpool layers with different sizes on input data, and concatenates their results at the end.

\subsection{Weld defect detection}
Due to the fast development of the machine learning and computer vision, lots of advanced algorithms are proposed to detect weld defects automatically. Andersen et al.~\cite{1990Artificial} were one of the earliest research teams that applied the artificial neural network in monitoring welding process. An improved multi-layer perceptron is proposed in~\cite{2016Automated}, and its detection accuracy reaches 88.6\%. Zhang et al.~\cite{2019Weld} designed an optical measurement system to capture images from three directions, then achieves weld detection using a simple CNN model. Two different deep convolutional neural networks~\cite{zhang2019weld} are trained on the same image set, and then they are transformed into a multi-model integrated frameworks to reduce the error rate. Sizyakin et al.~\cite{sizyakin2019automatic} integrated convolutional neural network and support vector machine, which improves positioning accuracy. An simplified YOLO v3 network is designed in~\cite{2020Automatic} through optimizing the loss function, and it obtains a great detection results. To achieve automatically feature extraction of laser welding images,~\cite{J2016Intelligent} designed the deep auto-encoding neural network which could extract salient, low-dimensional image features from high-dimensional laser welding images. In spite of success, the current weld defect detection methods do not consider the scale variation among different defect categories and the contextual information of image. In addition, the simple introduction of CNN is not efficient enough for industry as well.

%An incredible recognition framework requires an outstanding baseline, and advanced lightweight tools. To achieve this goal, we design our classification network based on ResNet~\cite{he2016deep} and the Conv method used in MobileNet~\cite{2018MobileNetV2}. Inspired by the fast speed of one-stage model, we propose our recognition network based on Tiny-YOLO v3~\cite{redmon2018yolov3}, which is an official compressed version of YOLO v3~\cite{redmon2018yolov3}. In addition, the Ghost Conv~\cite{han2020ghostnet} and feature map enhancement methods~\cite{hu2018squeeze}~\cite{bochkovskiy2020yolov4} are introduced to make model have a better performance.

\section{Method}

In this section, we introduce two core components, i.e., efficient feature extraction (EFE) module and reinforced multiscale feature (RMF) module in Section III-A and Section III-B, respectively. Furthermore, we design a weld defect detection network LF-YOLO based on the above modules in Section III-C.

\subsection{EFE module}

\begin{figure*}[!t]
  \centering
  \includegraphics[width=6.8in]{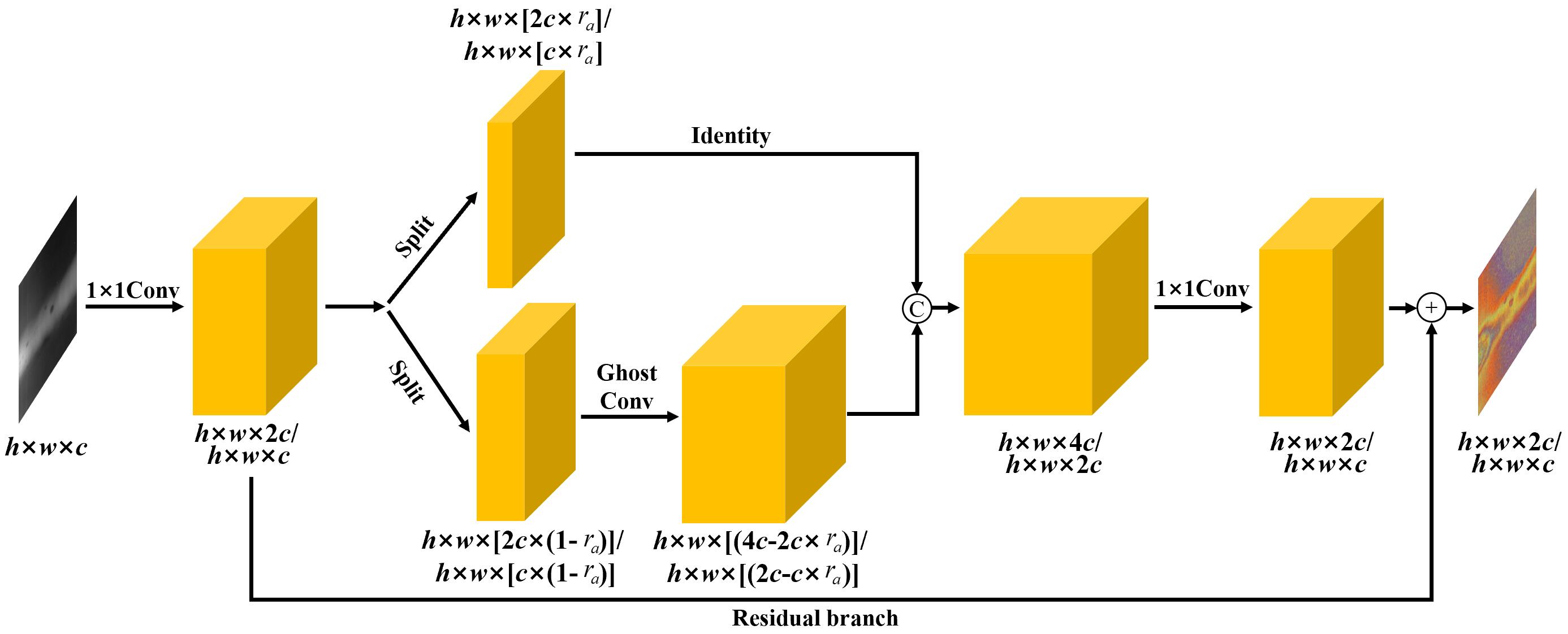}
  \caption{Illustration of our EFE module. It expands the weld feature in channel space to encode more implicit information. The idea of “split-transform-merge” is introduced to reuse the input, and Ghost Conv is used to reduce the module complexity.}
  \label{fig:EFE}
\end{figure*}

Feature extraction module is the basic block of deep learning network. An effective representation encoding enables model to better accomplish corresponding tasks. In addition, feature extraction operation is the main source of parameters and computation. Therefore, the weight of feature extraction module determines the weight of whole network. In this paper, we design an EFE module as shown in Fig.~\ref{fig:EFE}, which is lightweight but high-performance for representation learning of X-ray weld image.

EFE module firstly consists of a 1×1 convolution (Conv), and it increases the dimension of input feature maps to 2$c$ or maintains it as $c$, while the input data $I_{EFE} \in \mathbb{R}^{h \times w \times c}$. The choice depends on its location in network. Inspired by the inverted residual block in MobileNetV2~\cite{2018MobileNetV2}, EFE module maps the input data into a higher dimension space in the middle stage, because the expansion of feature space is beneficial to obtain more meaningful representation. But this mapping operation also would bring more memory and computation burden. MobileNetV2~\cite{howard2017mobilenets} solves this problem by using depthwise separable convolutions. In this paper, we employ a more wise strategy.
Following the idea of~\cite{xie2017aggregated}, we design the middle expansion structure based on “split-transform-merge” theory. After the first 1×1 Conv, feature maps are split into two branches, and split ratio \emph{$r_{a}$} is set as 0.25 in this paper. One of them is an identity branch, which does not utilize any operation on the data. Another branch is a dense block in~\cite{wang2020cspnet}, which is used to further extract features. However, dense block is complicated and high consumption due to its dense connection. To optimize the complexity, EFE module introduces Ghost Conv~\cite{han2020ghostnet}.

Ghost Conv can avoid redundant computation and Conv filters produced by similar intermediate feature maps, and achieves a great balance between accuracy and compression. We define the input feature map as $M \in \mathbb{R}^{h' \times w' \times c'},$ where $h'$, $w'$ and $c'$ are the height, width and the channel numbers of the input, respectively. We can formulate a conventional convolution process for generating feature maps $N$ as:
\begin{equation}
N = M \odot f,
\label{eq:conventional convolution}
\end{equation}
where $\odot$ represents the convolution operation, $f \in \mathbb{R}^{k \times k \times a \times  c'}$ is the convolution filters, and $k \times k$ is the kernel size of filters $f$. The output feature map $N \in \mathbb{R}^{h'' \times w'' \times a}$ has $a$ channels with $h''$ height and $w''$ width. To simplify the formulation, we omit the bias values that may be used.

But for Ghost Conv, it produces some intrinsic feature maps at first. Specifically, intrinsic feature maps $N^{\prime} \in \mathbb{R}^{h'' \times w'' \times b}$ are produced using conventional convolution:
\begin{equation}
N^{\prime}=M \odot f^{\prime},
\label{eq:intrinsic feature maps}
\end{equation}
where $f^{\prime} \in \mathbb{R}^{k \times k \times b \times c'}$ is the used convolution filters, and $b \leq a$. The values of height $h''$ and width $w''$ are remained unchanged to make the spatial size consistent compared with Eq.~\ref{eq:conventional convolution}. To obtain the required feature map whose channel number is $a$, Ghost Conv utilizes cheap linear operations on each intrinsic feature to generate needed $s$ ghost features according to the following function:

\begin{equation}
n_{i j}=\mathfrak{L}_{i, j}\left(n_{i}^{\prime}\right), \quad \forall i=1, 2, \ldots, b, \quad j=1, 2, \ldots, s,
\label{eq:cheap linear operations}
\end{equation}
where $n_{i}^{\prime}$ is the $i$-th intrinsic feature map of $N^{\prime}$, $\mathfrak{L}_{i, j}$ is the linear operation that produces the $j$-th ghost feature map $n_{i j}$. Finally, we can obtain $b=a \cdot s$ feature maps $N=\left[n_{11}, n_{12}, \cdots, n_{b s}\right]$ as the output.

The results of two branch are concatenated at last, and the output dimension of the expansion operation is set to twice the input. The middle expansion operation is a well-designed block. Firstly, split operation reuses the input features through retaining a portion of the channels, which enhances the information flow between different layers. Expansion operation occupies the main memory and computation consumption theoretically, because it produces higher dimension. However, the introduction of Ghost Conv alleviates the burden to a great extent.

At the tail of EFE module, the second 1×1 Conv is used to compress the number of channels back to 2$c$/$c$. Finally, the input of expansion operation and the output of second 1×1 Conv are added element-wise by a residual branch. YOLO v3~\cite{redmon2018yolov3} also uses residual block to extract features, we show the complexity comparison between our EFE module and residual block used in~\cite{redmon2018yolov3}. The stride of first Conv in~\cite{redmon2018yolov3} is set as 2 to downsample, and we change it to 1 same as our RMF module for fairness. Table~\ref{tab:residual-comparison} shows the parameters (Params) and floating-point of operations (FLOPs) of two structures while input is 208$\times$208$\times$128 and output is 208$\times$208$\times$256. Compared with the conventional residual block, our EFE module greatly decreases the consumption of feature extraction.

\begin{table}[!t]
	\renewcommand{\arraystretch}{1.6}
	\renewcommand\tabcolsep{2pt}
	\caption{Complexity comparison between EFE module and residual block in YOLO v3}
	\centering
	\label{tab:residual-comparison}
	\begin{tabular}{cc|cc}
		\toprule
		&                                            & Params(M)      & FLOPs(G)        \\
		\midrule
        & EFE module                                 & \textbf{0.2}   & \textbf{9.1}                \\
        \hline
        & Residual block in YOLO v3 & 0.6            & 27.0                \\
		\bottomrule
	\end{tabular}
\end{table}

\subsection{RMF module}

Scale problem is a classical research topic for CNN, because it is not robust enough for the sizes of objects. Especially when the sizes of objects vary greatly, the plain topology model will encounter an awful performance. To obtain more effective feature maps of X-ray weld image through multi-scale strategy, we design a RMF module combining the parameter-based and parameter-free methods.

RMF module is a hierarchical structure for obtaining multi-scale contextual information, and its framework is shown in Fig.~\ref{fig:RMF}. It firstly leverages spatial pyramid pooling~\cite{bochkovskiy2020yolov4}, which utilizes multiple maxpool operations with different sizes on input feature map. There are not any parameters introduced in this stage, hence we regard it as parameter-free. Parameter-free method makes the most of existing data, but not generating new information in a sense. It enables maximum exploitation of the explicit information from existing feature maps and integrates them effectively. Each branch of spatial pyramid pooling is connected with another further hierarchy, and it is achieved through an improved dilated convolution proposed in this paper, namely Ghost-Dilated Convolution (GDConv).

\begin{figure*}[!t]
  \centering
  \includegraphics[width=6.0in]{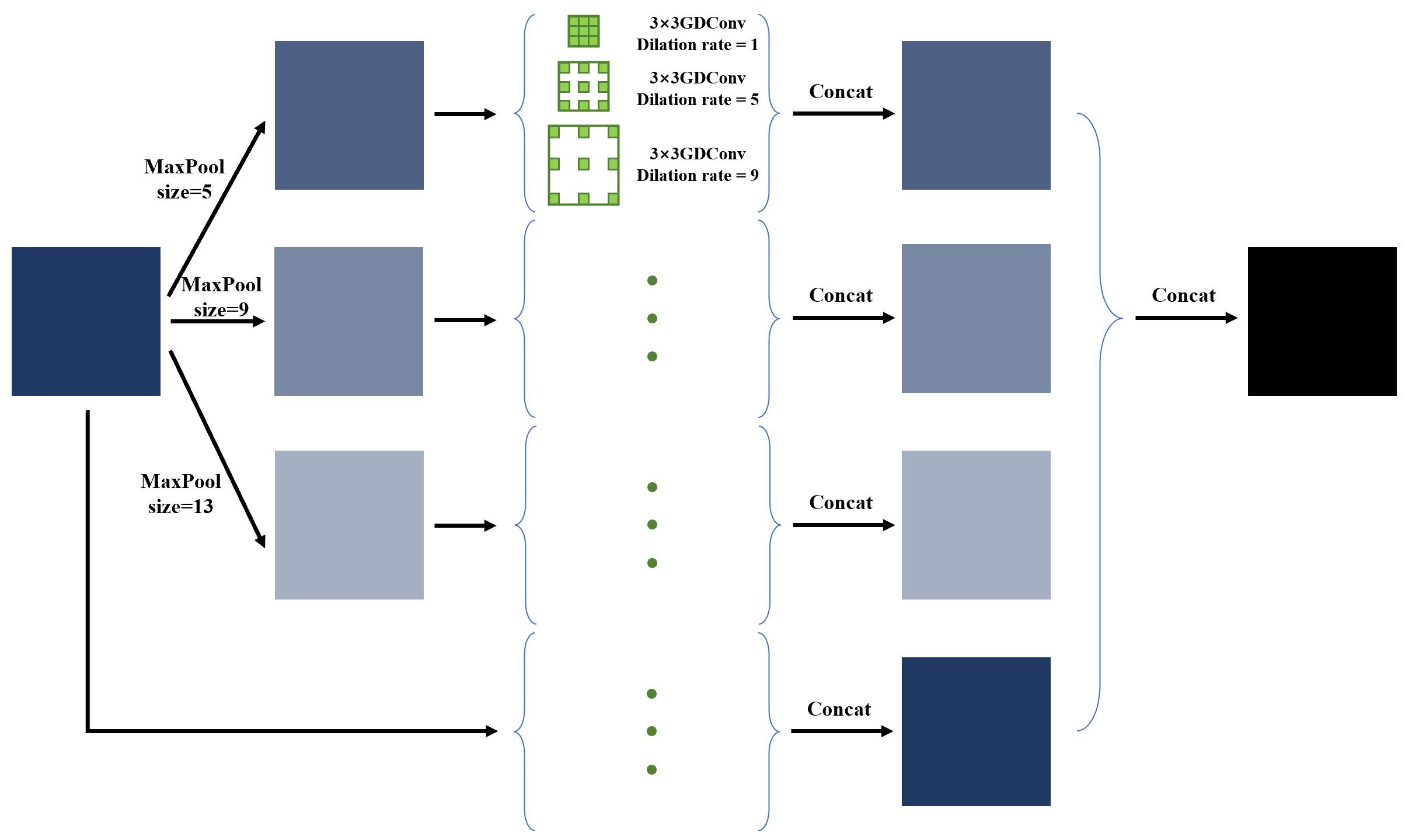}
  \caption{Illustration of our RMF module. The whole hierarchical structure contains a base multi-scale utilization operated by different maxpool (parameter-free), and a further stage to find underlying multi-scale feature achieved by novel GDConv (parameter-based). The results obtained by each stage will be combined and sent to the latter layer for a finer prediction.}
  \label{fig:EME}
\end{figure*}

\begin{figure}[!t]
  \centering
  \includegraphics[width=1.5in]{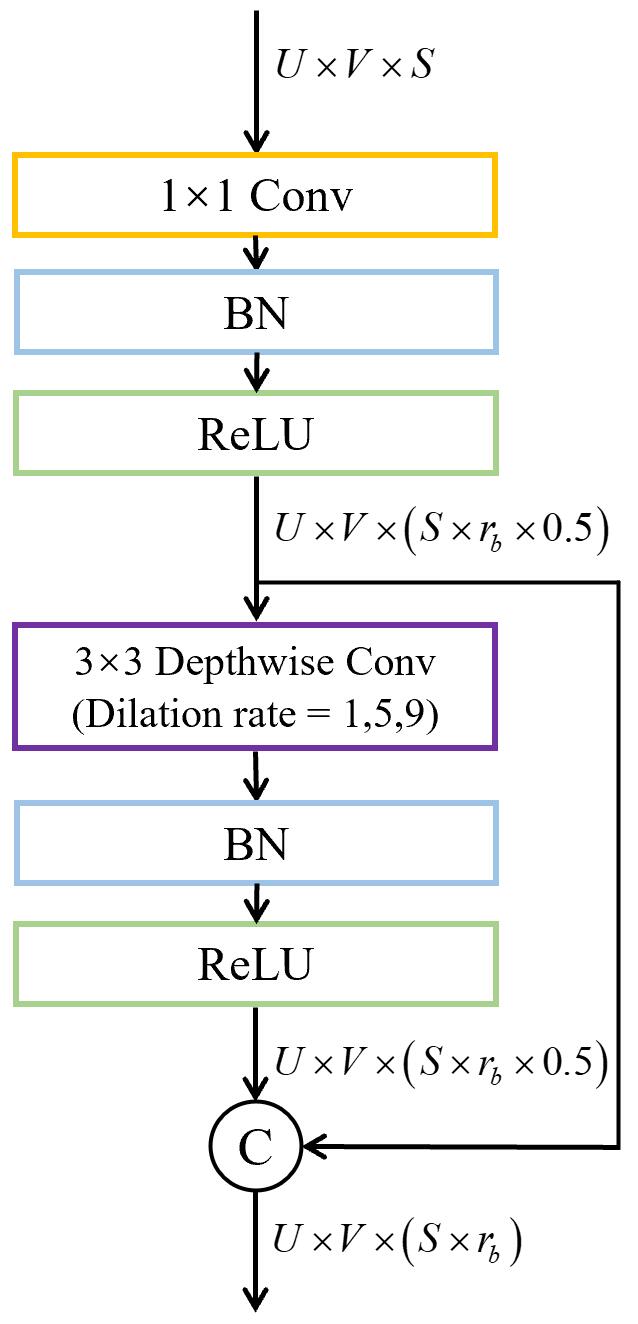}
  \caption{GDConv. It retains the structure of Ghost Conv, but implements depthwise Conv with different dilation rates. GDConv can capture cues from various receptive fields with lightweight design.}
  \label{fig:GDConv}
\end{figure}

Dilated convolution can enhance the ability to extract underlying information through changing the receptive field~\cite{yu2015multi}. However, dilated convolution can be regard as a variant of normal convolution without any lightweight improvement. When the layer is deeper, the high dimension burden would be aggravated. If we use dilated convolution directly at the tail of backbone, it would be expensive on storage and computation. To address this problem, GDConv achieves dilation process based on a lighter form. Specifically, we retain the structure of original Ghost Conv but operate depthwise Conv with dilation version, and its inner detail is shown in Fig.~\ref{fig:GDConv}. GDConv is the core ingredient for RMF module to learn implicit information through parameters of convolution kernels. Three GDConvs form the elements of a hierarchy group, and their dilation rates are set as 1, 5, 9 respectively. Note that when dilation rate is 1, it is equivalent to normal Ghost Conv, and the new features from different dilation branches would be concatenated. Furthermore, four arteries would converge at last through concatenation as well. RMF can be formulated as follow:

\begin{equation}
f_{ks} = F(Max_{ks}(I_{RMF}))
%O = Con\{Con\{GD(Max(I)_{p=1}^{P})_{q=1}^{Q}\}\},
\label{eq:EME-output}
\end{equation}

\begin{equation}
O_{RMF} = \oplus\{\oplus\{f_{1}, \cdots, f_{g}\}_{1}, \cdots, \oplus\{f_{1}, \cdots, f_{g}\}_{h}\},
%O = Con\{Con\{GD(Max(I)_{p=1}^{P})_{q=1}^{Q}\}\},
\label{eq:EME-output}
\end{equation}
where $I_{RMF}$ and $O_{RMF}$ represent the input and output of RMF module, respectively. $Max_{ks}(\cdot)$ is the maxpool operation with kernel size $ks$, and $F(\cdot)$ represents the learnable GDConv layer, while $f_{ks}$ is corresponding learned feature. $\oplus$ is concatenation operation, while $g$ and $h$ are the numbers of GDConv and maxpool, and they are set as 3 and 4 in our design.

The main idea of RMF module is to integrate explicit (produced by parameter-free method) and implicit (produced by parameter-based method) multi-scale information of weld image. As the former stage of whole hierarchy, maxpool operations generate a base pyramid, which has contained contextual information of different scale at first. Based on each level of the basic pyramid, new underlying features are encoded and generated from different receptive field scales. In other words, the parameter-free method provides a multi-scale base through optimizing existing feature maps, and parameter-based method exploits new multi-scale data based on the former. Hence, the base and expansion pyramid of hierarchy have a superposition effect and enhance the ability to better develop effective representation. To alleviate dimension stress, $r_{b}$ is set as 0.5, and each GDConv would reduce the channels of input by half. Finally, the output channel numbers of RMF is six times as that of input.

\subsection{The architecture of LF-YOLO}
\begin{figure}[!t]
  \centering
  \includegraphics[width=2.4in]{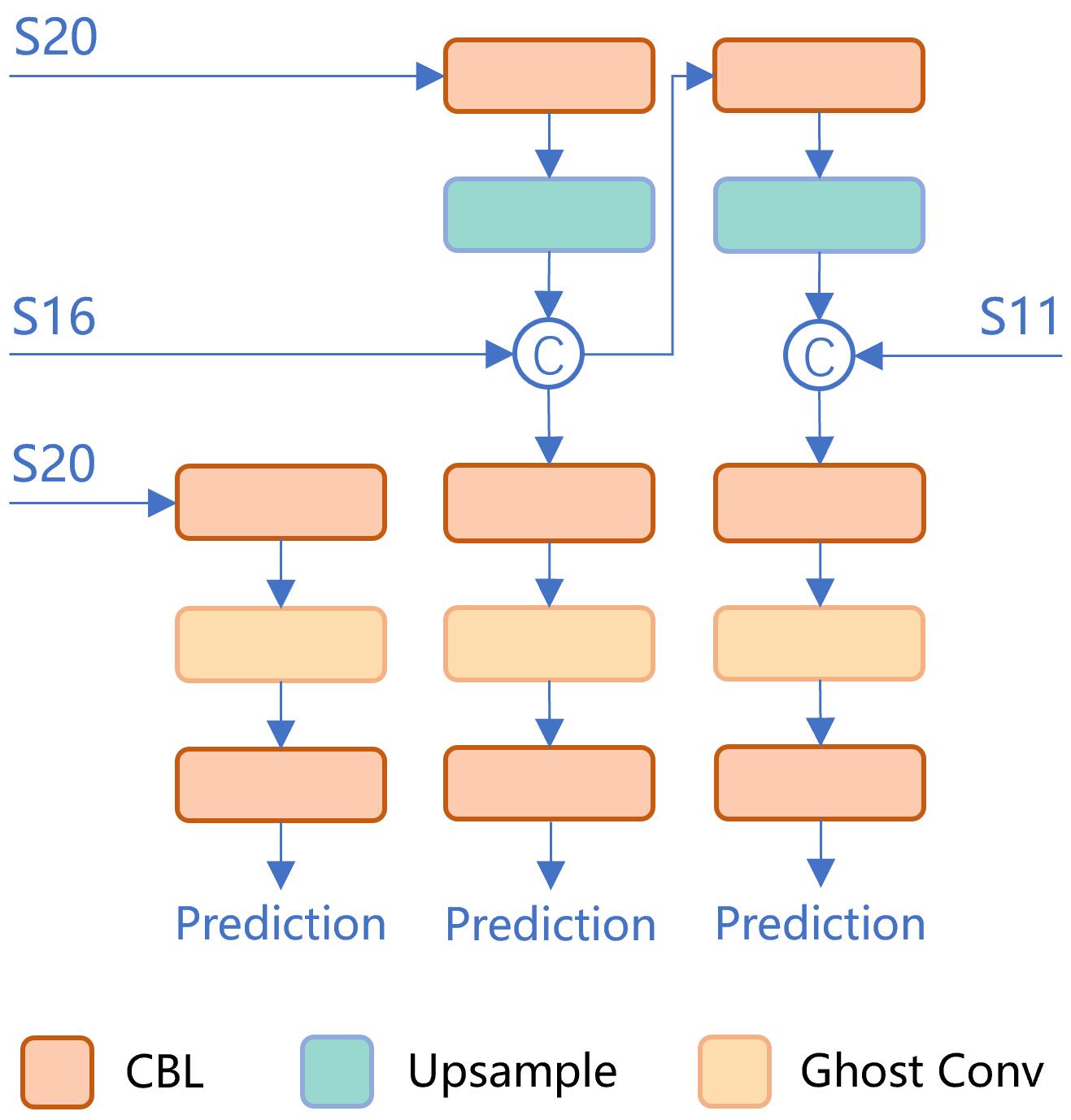}
  \caption{The prediction heads of LF-YOLO.}
  \label{fig:prediction}
\end{figure}

Based on the above modules, we design a new detection network LF-YOLO for weld defect, and its backbone is shown in Table~\ref{tab:backbone}. Original input weld image is processed by Conv-BatchNorm-LeakyReLU (CBL) for initial feature extraction, and the dimension of original input is $H\times W \times 3$. Furthermore, network is constructed with maxpool and EFE, while the former one is used to decrease the resolution of feature map and the latter one is introduced for deeper representation learning. At the tail of backbone, the obtained feature would be enhanced via RMF module. LF-YOLO inherits the structure of feature pyramid network in YOLO, and three detection heads are designed for different size objects. We use CBL with size of 1×1 and Ghost Conv to further adjust feature extracted from backbone according to the size of each prediction output as shown in Fig.~\ref{fig:prediction}.

For objectness and classfication prediction, we compute the distance between label $x_{n}$ and prediction result $y_{n}$ through binary cross entropy loss, which can be defined as follow:

\begin{equation}
\mathcal{L}_{obj/cls} = y_{n} \cdot \log x_{n}+\left(1-y_{n}\right) \cdot \log \left(1-x_{n}\right).
\label{eq:BCELoss}
\end{equation}

Intersection over union (IOU) loss is introduced to regress the bounding box, and it is defined as:

\begin{equation}
\mathcal{L}_{box} = 1 - IOU^{gt}_{pre},
\label{eq:BCELoss}
\end{equation}
where $IOU^{gt}_{pre}$ represents the value of IOU between ground truth and prediction box.

Finally, the total loss of LF-YOLO is calculated by combining above three parts:
\begin{equation}
\mathcal{L}_{total} = \mathcal{L}_{obj} +\mathcal{L}_{cls} + \mathcal{L}_{box}.
\label{eq:Loss}
\end{equation}

\begin{table}[!t]
\renewcommand{\arraystretch}{1.5}
\renewcommand\tabcolsep{10pt}
\centering
\caption{The backbone network of LF-YOLO}
\begin{tabular}{cccccc}
\toprule
Layer name & Type     & Output size \\
\hline
S1         & CBL      &$H\times W\times 16$   \\
\hline
S2         & MaxPool  &$H/2\times W\/2\times C$   \\
\hline
S3         & EFE      &$H/2\times W/2\times 2C$  \\
\hline
S4         & MaxPool  &$H/4\times W/4\times 2C$  \\
\hline
S5         & EFE      &$H/4\times W/4\times 4C$  \\
\hline
S6         & EFE      &$H/4\times W/4\times 4C$  \\
\hline
S7         & MaxPool  &$H/8\times W/8\times 4C$  \\
\hline
S8         & EFE      &$H/8\times W/8\times 8C$  \\
\hline
S9         & EFE      &$H/8\times W/8\times 8C$  \\
\hline
S10        & EFE      &$H/8\times W/8\times 8C$  \\
\hline
S11        & EFE      &$H/8\times W/8\times 8C$  \\
\hline
S12        & MaxPool  &$H/16\times W/16\times 8C$  \\
\hline
S13        & EFE      &$H/16\times W/16\times 16C$  \\
\hline
S14        & EFE      &$H/16\times W/16\times 16C$  \\
\hline
S15        & EFE      &$H/16\times W/16\times 16C$  \\
\hline
S16        & EFE      &$H/16\times W/16\times 16C$  \\
\hline
S17        & MaxPool  &$H/32\times W/32\times 16C$  \\
\hline
S18        & EFE      &$H/32\times W/32\times 32C$  \\
\hline
S19        & EFE      &$H/32\times W/32\times 32C$  \\
\hline
S20        & RMF      &$H/32\times W/32\times 196C$  \\
\bottomrule
\end{tabular}
\label{tab:backbone}
\end{table}

\section{Experiments}
In this section, we firstly introduce the experimental setup including dataset, implementation details and evaluation metrics. Then, we compare our network with other state-of-the-art methods on our weld defect dataset. Ablation studies are used to prove the contributions of EFE and RMF module. Finally, to prove the versatility of our model, we present the quantitative comparisons with classical models on public dataset.

\subsection{Experimental Setup}

\label{sec:robustness_anal}
\textbf{Datasets.}  We build a weld defect dataset including 2449 digital X-ray images, which are from the actual special equipment companies. All images have been annotated carefully by professionals, and there are three defect categories containing blow hole, incomplete penetration and crack. We randomly split them to train set and test set in a 9:1 ratio. To prove the versatility of our model, we introduce the MS COCO~\cite{lin2014microsoft} to evaluate our network. MS COCO dataset has 118k training images with 80 categories and we report detection results on its test set.

\textbf{Implementation Details.} We conduct all experiments on a i7-8700K CPU and a single NVIDIA GeForce GTX2070 GPU. All models are based on deep learning framework PyTorch. We choose stochastic gradient descent optimizer with 0.9 momentum and 0.0005 weight decay. The initial learning rate and total epochs are set as 0.01 and 500 for weld defect dataset, while 0.01 and 300 for MS COCO. The images in weld dataset are resized to 320$\times$320 for training and testing, while images in MS COCO are resized to 640$\times$640. The state-of-the-art methods are trained and tested based on MMdetection~\cite{chen2019mmdetection} which is an open source 2D detection toolbox, and all related hyper-parameters are adjusted to the optimal. We obtain different versions LF-YOLO$\times$$n$ through changing the width $C$ defined in Table~\ref{tab:backbone}. $n$ is proportional to $C$ while $n$=1, $C$=32.

\textbf{Evaluation Metrics.} We adopt mAP$_{50}$, FLOPs, Params and frame per second (FPS) as evaluation metrics to evaluate proposed network comprehensively. mAP$_{50}$ is a common metric to evaluate comprehensive ability over all classes. It is simply the average AP$_{50}$ over all classes, and AP$_{50}$ can be computed as the area of Precision $\times$ Recall curve with axles while the IOU threshold is set as 0.5~\cite{padilla2020survey}.

Moreover, to compare the computation complexity of different networks, time complexity (FLOPs) and space complexity (Params) are chosen to show the weights of different methods. In addition, we use FPS to show the speed during inference stage, and the results of FPS are the average of 245 testing images of weld defect dataset.

\subsection{Comparisons with state-of-the-art models}

\begin{table}[!t]
	\renewcommand{\arraystretch}{1.6}
	\renewcommand\tabcolsep{2.5pt}
	\caption{The weld defect detection results in comparison with state-of-the-art methods}
	\centering
	\label{tab:weld results}
	\begin{tabular}{l|c|cccccccc}
		\toprule
		Methods             & Backbone   & mAP$_{50}$       & FPS           & FLOPs(G) & Params(M)    \\
		\midrule
        Cascasde-RCNN       & ResNet50   & 90.0             & 10.2          & 243.2    & 68.9 \\
        Cascasde-RCNN       & ResNet101  & 90.7             & 8.0           & 323.1    & 87.9 \\
        Faster-RCNN         & ResNet50   & 90.1             & 13.7          & 215.4    & 41.1 \\
        Faster-RCNN         & ResNet101  & 92.2             & 10.3          & 295.3    & 60.1 \\
        Dynamic-RCNN        & ResNet50   & 90.3             & 13.4          & 215.4    & 41.1 \\
        \hline
        RetinaNet           & ResNet50   & 80.0             & 15.0          & 205.2    & 36.2 \\
        VFNet               & ResNet50   & 87.0             & 13.5          & 197.8    & 32.5 \\
        VFNet               & ResNet101  & 87.2             & 10.2          & 277.7    & 51.5 \\
        Reppoints           & ResNet50   & 82.7             & 13.9          & 199.0    & 36.6 \\
        SSD300              & VGGNet     & 88.1             & 43.8          & 30.6     & 24.0 \\
        YOLO v3             & Darknet52  & 91.0             & 38.6          & 33.1     & 62.0 \\
        \hline
        SSD300              & MobileNetV2  & 82.3          & 57.3          & \textbf{0.7} & 3.1  \\
        YOLO v3             & MobileNetV2  & 90.2          & 71.0          & 1.6      & 3.7  \\
        \hline
        LF-YOLO$\times0.5$  & $-$        & 90.7             & \textbf{72.8} & 1.1      & \textbf{1.8}  \\
        LF-YOLO             & $-$        & \textbf{92.9}    & 61.5          & 4.0      & 7.3  \\
		\bottomrule
	\end{tabular}
\end{table}

\begin{figure*}[!t]
  \centering
  \includegraphics[width=5.8in]{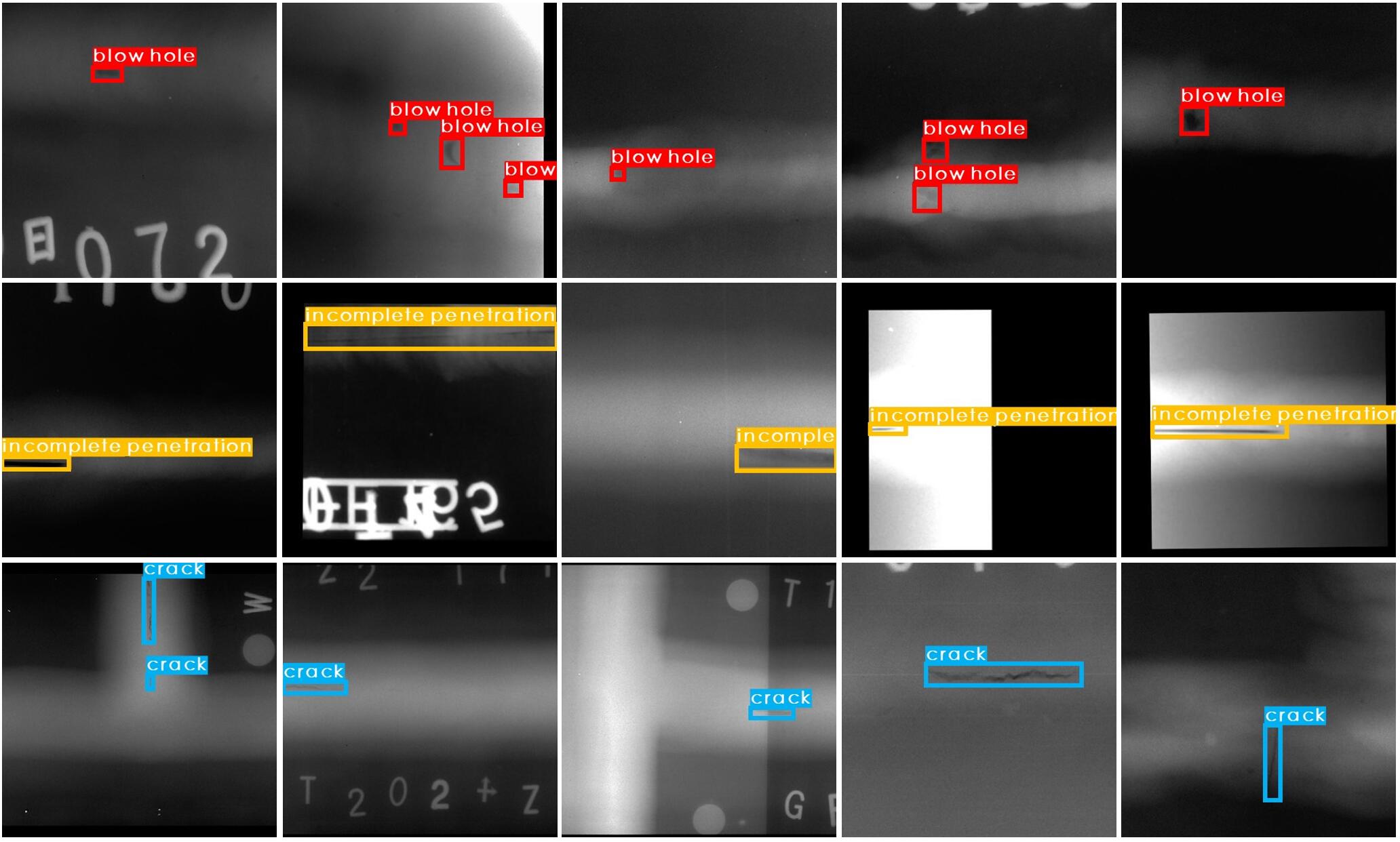}
  \caption{The samples of detection results of weld defects. Our LF-YOLO has an outstanding prediction on all defect classes, even the defects are small and dim. It shows that LF-YOLO is capable of achieving satisfactory prediction by combing powerful contextual information and local textural features.}
  \label{fig:weld_results}
\end{figure*}

To validate the weld defect detection performance of our LF-YOLO, we compare our method with many state-of-the-art models, including two-stage networks Cascasde-RCNN~\cite{cai2018cascade}, Faster-RCNN~\cite{ren2015faster} and Dynamic-RCNN~\cite{DynamicRCNN}, one-stage networks RetinaNet~\cite{lin2017focal}, VFNet~\cite{zhang2020varifocalnet}, Reppoints~\cite{yang2019reppoints}, SSD300~\cite{liu2016ssd} and YOLO v3~\cite{redmon2018yolov3}. In addition, we replace the default backbone of SSD300 and YOLO v3 with lightweight backbone MobileNetV2~\cite{2018MobileNetV2}, and all  compared models are trained and tested based on MMDetection~\cite{chen2019mmdetection}. Table~\ref{tab:weld results} shows the quantitative comparison results on our weld defect dataset. Our method LF-YOLO achieves 92.9 mAP$_{50}$ which outperforms all other methods, and the complexity of it is much lower than all classical one-stage and two-stage models with only 7.3M Params and 4.0G FLOPs. MobileNet v2-SSD300 has the lowest FLOPs, but its detection performance is far from satisfactory with only 82.3 mAP$_{50}$. Our LF-YOLO$\times0.5$ achieves the best results on detection speed and Params, and its detection performance is superior compared with all lightweight network, most of one-stage models and some two-stage networks, while only worse than Faster-RCNN with ResNet50 and vanilla YOLO v3. However, the efficiency of LF-YOLO$\times0.5$ is much better than that of these methods. In addition, we report the results of each defect class as shown in Table~\ref{tab:each-class-results}, and our LF-YOLO has an outstanding prediction on all defect classes. The result of blow hole is worse than other two defects, because the distribution of it is dense, and extremely small labels appear sometimes. Some detection samples are given in Fig.~\ref{fig:weld_results}.

\begin{table}[!t]
	\renewcommand{\arraystretch}{1.6}
	\renewcommand\tabcolsep{2pt}
	\caption{mAP$_{50}$ on each class of weld defect predicted by LF-YOLO}
	\centering
	\label{tab:each-class-results}
	\begin{tabular}{cc|c|c}
		\toprule
		& Blow hole      & Incomplete peneration   & Crack\\
		\midrule
        & 84.2           & 96.0                    & 98.6 \\
		\bottomrule
	\end{tabular}
\end{table}

\subsection{Ablation studies}
We design ablation studies for EFE and RMF modules based on LF-YOLO$\times0.5$, which can prove the theoretical basis of this paper. In EFE module, we define STM is short for split-transform-merge design, and SPP for spatial pyramid pooling structure. From Table~\ref{tab:ablation-in-EFE}, we can conclude that the whole EFE module reaches the best balance. While replacing the Ghost Conv with normal Conv, it obtains 90.8 mAP$_{50}$ which is only 0.1 point higher than intact module. But under this design, its FLOPs and Params are 2.7$\times$ and 2.3$\times$ than LF-YOLO$\times0.5$. Compared with other combinations, our LF-YOLO$\times0.5$ outperforms on performance, computation and storage.

RMF is consisted of SPP and the proposed GDConv. As shown in Table~\ref{tab:ablation-in-RMF}, satisfactory results can only be obtained when they are used together. Removing SPP means that the input feature would be processed by several GDConv directly without parameter-free operation, while removing GDConv means that the parameter-based stage is deleted. The above two cases decrease the capacity of model greatly. When both of them are removed, the degeneration is the worst with only 86.3 mAP$_{50}$. Our design has the best detection results with slightly higher model complexity.

\begin{table}[!t]
	\renewcommand{\arraystretch}{1.6}
	\renewcommand\tabcolsep{4pt}
	\caption{Ablation study of EFE module}
	\centering
	\label{tab:ablation-in-EFE}
	\begin{tabular}{cc|ccc}
		\toprule
		STM        & Ghost Conv       & mAP$_{50}$     & FLOPs(G)      & Params(M)    \\
		\midrule
                   &                  & 90.3           & 4.1           & 5.6 \\
                   & \checkmark       & 89.4           & 1.2           & 1.9 \\
        \checkmark &                  & \textbf{90.8}  & 3.0           & 4.2 \\
        \checkmark & \checkmark       & 90.7           & \textbf{1.1}  & \textbf{1.8} \\
		\bottomrule
	\end{tabular}
\end{table}

\begin{table}[!t]
	\renewcommand{\arraystretch}{1.6}
	\renewcommand\tabcolsep{5pt}
	\caption{Ablation study of RMF module}
	\centering
	\label{tab:ablation-in-RMF}
	\begin{tabular}{cc|ccc}
		\toprule
		SPP        & GDConv     & mAP$_{50}$     & FLOPs(G)          & Params(M)    \\
		\midrule
                   &            & 86.3           & \textbf{1.0}      & \textbf{1.1} \\
                   & \checkmark & 87.9           & \textbf{1.0}      & \textbf{1.1} \\
        \checkmark &            & 87.8           & \textbf{1.0}      & 1.4 \\
        \checkmark & \checkmark & \textbf{90.7}  & 1.1               & 1.8 \\
		\bottomrule
	\end{tabular}
\end{table}

\subsection{Versatility validation on public dataset}

To prove the detection versatility of our method, we test it on convincing public dataset (e.g., MS COCO). We train our models on COCO trainval35k~\cite{lin2014microsoft} and evaluate them on test-dev2017 by submitting the results to COCO server. We compare our models with classical one-stage networks SSD~\cite{liu2016ssd} and YOLO v3~\cite{redmon2018yolov3}, and two-stage networks Faster-RCNN~\cite{ren2015faster} and R-FCN~\cite{cai2018cascade}. The results of compared methods are from the cited references of table. Our LF-YOLO$\times0.75$ reaches 43.9 mAP$_{50}$ which is higher than SSD300 and same as Faster-RCNN, and the computation of it is only 28.7\% and 25.9\% of that of them. LF-YOLO$\times1.25$ achieves better performance compared with most of models, while using lower FLOPs and fewer Params. R-FCN reaches the best mAP$_{50}$ results with 52.9, which is 2.6 points higher than LF-YOLO$\times1.25$ , but its weight is unacceptable for industry. Hence, our method still has a satisfactory balance between performance and lightweight when tested on public dataset.

\begin{table}[!t]
	\renewcommand{\arraystretch}{1.6}
	\renewcommand\tabcolsep{3pt}
	\caption{COCO test-dev detection results}
	\centering
	\label{tab:}
	\begin{tabular}{l|c|ccccccc}
		\toprule
		Methods                         & Backbone & mAP$_{50}$ & FLOPs(G) & Params(M)   \\
		\midrule
        SSD300~\cite{liu2016ssd}        & VGGNet    & 41.2       & 35.2     & 34.3           \\
        SSD512~\cite{liu2016ssd}        & VGGNet    & 46.5       & 99.5     & 34.3           \\
        YOLO v3~\cite{redmon2018yolov3} & Darknet52 & 51.5       & 39.0   & 61.9           \\
        Faster R-CNN~\cite{cai2018cascade}& VGGNet    & 43.9       & $-$    & 278.0        \\
        R-FCN~\cite{cai2018cascade}     & ResNet50  & 49.0       & $-$      & 133.0        \\
        R-FCN~\cite{cai2018cascade}     & ResNet101 & \textbf{52.9}  & $-$      & 206.0        \\
        \hline
        LF-YOLO$\times0.75$             & $-$       & 43.9       & \textbf{10.1}     & \textbf{4.2}    \\
        LF-YOLO                         & $-$       & 47.8       & 17.1     & 7.4            \\
        LF-YOLO$\times1.25$             & $-$       & 50.3       & 26.0     & 11.5            \\
		\bottomrule
	\end{tabular}
\end{table}

\subsection{Visualization of intermediate feature maps}
\begin{figure*}[!t]
  \centering
  \includegraphics[width=6in]{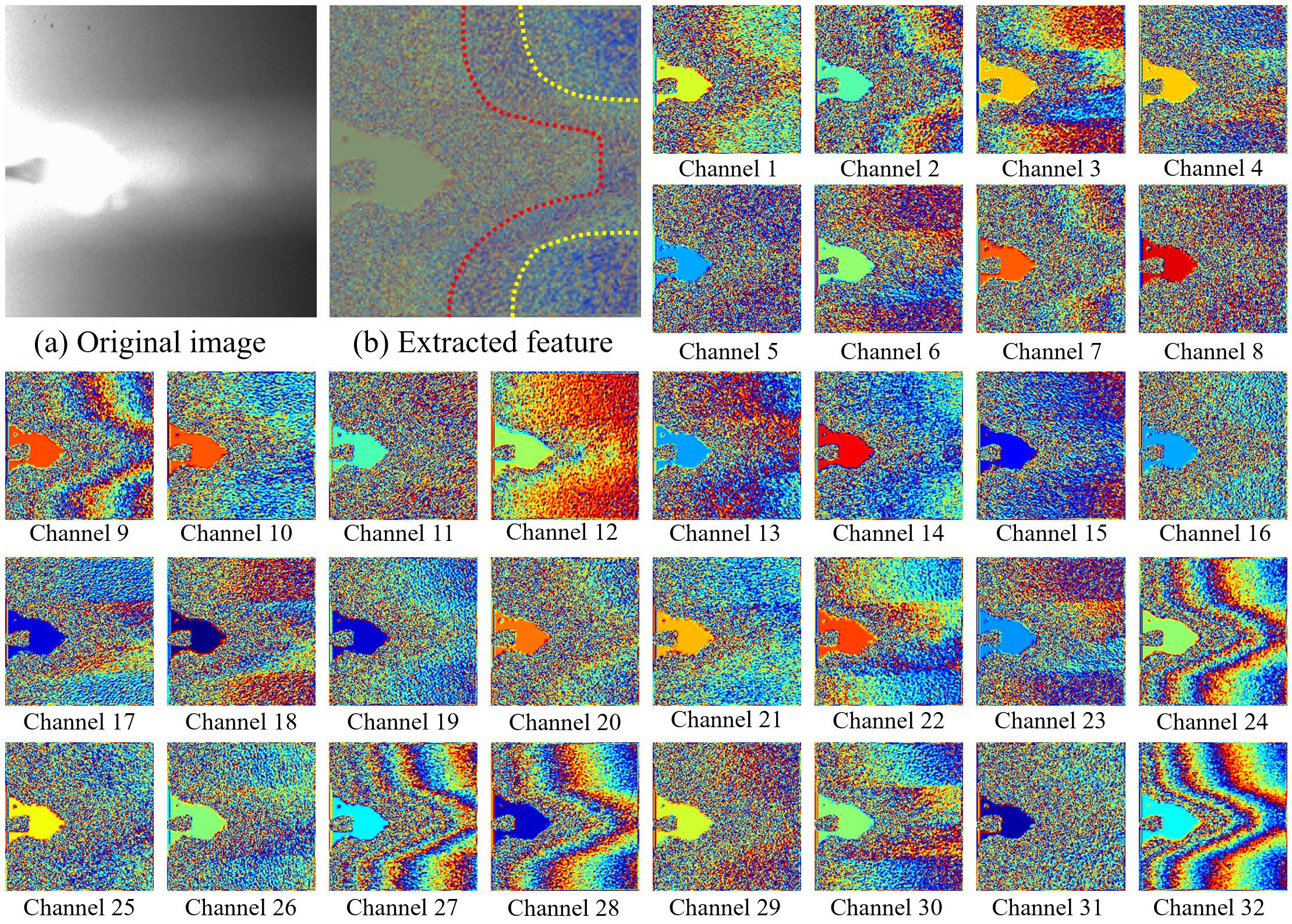}
  \caption{The visualization of feature maps produced by first EFE module on each channel. The extracted feature map owns superior region division, and fine-grained information which can be shown by the dotted line with red and yellow. It contributes to the fact that features learned through EFE module are effective and useful, which can be concluded by observing each channel.}
  \label{fig:visualization}
\end{figure*}

To analyse the effectiveness of our network, especially the feature extraction block, we visualize the feature maps produced by the first EFE module of LF-YOLO as shown in Fig.~\ref{fig:visualization}. From Fig.~\ref{fig:visualization} (b), we can observe that the feature extracted by EFE is satisfactory in distinguishing different regions. Furthermore, it represents the fine-grained information caused by the change of brightness, which is shown through dotted line with red and yellow. Even this variation is hard for human to notice from Fig.~\ref{fig:visualization} (a). For a more detailed analysis, we show the feature of each channel. There are some channels that pay more attention to the underlying texture information, such as channel 9, 24 and 32. Some channels more purely focus on region segmentation, such as channel 8, 11, 31. Hence, EFE module can encode sufficient information of X-ray weld image including region distribution and geometrical cues, and provide reliable features for final estimation.

\section{Conclusion}
In this paper, we propose a highly effective EFE module as the basic feature extraction block, and it can encode sufficient information of X-ray weld image with low consumption. In addition, a novel RMF module is designed to utilize and explore multi-scale feature. The parameter-free stage contributes to a basis containing existing multi-scale information, and parameter-based stage further learn implicit feature among different receptive fields. Finally, the results from two stages will be combined for more powerful representation. Based on the two above modules, we design a LF-YOLO model for weld defect detection. LF-YOLO obtains 92.9 mAP$_{50}$ with only 4.0G FLOPs, 7.3M Params and 61.5 FPS on weld defect dataset, and achieves superior balance between speed and performance. On public object dataset MS COCO, our LF-YOLO also is competitive, reaching 50.3 mAP$_{50}$ by $\times1.25$ version. In the future, we will further optimize the algorithm, and apply it on embedded platform. Furthermore, we will collect more data and improve the detection performance of network, especially for blow hole.

\bibliographystyle{IEEEtran}
\bibliography{References}

% that's all folks
\end{document}